\documentclass[twocolumn]{article}

\usepackage{titlesec}
\usepackage{cite}
\usepackage{url}
\usepackage[utf8]{inputenc}
\usepackage{array}
\usepackage{color}
\usepackage{colortbl}
\usepackage{booktabs}
\usepackage{lmodern} % Modern fonts

\titleformat{\section}
  {\normalfont\Large\bfseries}{\thesection}{1em}{}
\titlespacing*{\section}
  {0pt}{*1.0}{*0.5} % {left}{before-sep}{after-sep}

\titleformat{\subsection}
  {\normalfont\large\bfseries}{\thesubsection}{1em}{}
\titlespacing*{\subsection}
  {0pt}{*0.8}{*0.4}

\titleformat{\subsubsection}
  {\normalfont\normalsize\bfseries}{\thesubsubsection}{1em}{}
\titlespacing*{\subsubsection}
  {0pt}{*0.6}{*0.3}

\usepackage{lipsum} % for generating dummy text
\usepackage{booktabs} % For formal tables
\usepackage[table,xcdraw]{xcolor} % for table row coloring
\usepackage{tabularx}
\usepackage{pdflscape}
\usepackage{array} % For table wrapping
\usepackage{float}
\usepackage{tikz}
\usepackage{multirow}
\usepackage{adjustbox}
\usepackage{siunitx} % For aligning numbers by decimal point
\usepackage{makecell}
\usepackage{pifont} % For symbols
\usepackage{amssymb}
\usepackage{geometry}
\usepackage{graphicx}
\usepackage{multicol}
\usepackage{comment}
\usepackage{colortbl}

\definecolor{lowrisk}{RGB}{144,238,144} % Light green
\definecolor{moderaterisk}{RGB}{255,255,102} % Light yellow
\definecolor{highrisk}{RGB}{255,150,120}  % Lighter red
\definecolor{criticalrisk}{RGB}{255,120,120} % Lighter darker red
\definecolor{chatgpt4}{RGB}{230,240,255}
\definecolor{chatgpt4o}{RGB}{235,245,255}
\definecolor{chatgpt4mini}{RGB}{240,250,255}
\definecolor{claude35}{RGB}{245,255,245}
\definecolor{gemini}{RGB}{255,245,245}
\definecolor{mistral}{RGB}{255,250,240}
\definecolor{Gray}{gray}{0.9}
\definecolor{chatgpt4}{RGB}{52,168,83}
\definecolor{chatgpt4o}{RGB}{66,133,244}
\definecolor{chatgpt4mini}{RGB}{251,188,5}
\definecolor{claude35}{RGB}{234,67,53}
\definecolor{claudehaiku}{RGB}{255,109,1}
\definecolor{gemini}{RGB}{103,58,183}
\definecolor{geminiadvanced}{RGB}{63,81,181}
\definecolor{mistral}{RGB}{0,150,136}
\definecolor{lightblue}{RGB}{200,220,245}
\definecolor{lightyellow}{RGB}{255,255,224}
\definecolor{lightorange}{RGB}{255,229,180}
\definecolor{lightred}{RGB}{255,204,203}
\definecolor{lightgreen}{RGB}{144,238,144}
\definecolor{lightgray}{RGB}{245,245,245}
\definecolor{accessiblered}{RGB}{255,153,153}
\definecolor{accessibleorange}{RGB}{255,204,153}
\definecolor{accessibleyellow}{RGB}{255,255,153}
\definecolor{accessiblegreen}{RGB}{153,255,153}
\definecolor{accessiblelightblue}{RGB}{135,206,250}
\begin{document}

\title{Advice for Diabetes Self-Management by ChatGPT Models: Challenges and Recommendations}
\author{
  Waqar Hussain, CSIRO's Data61, Australia\thanks{waqar.hussain@data61.csiro.au} \and
  John Grundy, Monash University, Australia\thanks{john.grundy@monash.edu}
}
\date{}

\maketitle

\textbf{Abstract: }

Given their ability for advanced reasoning, extensive contextual understanding, and robust question-answering abilities, large language models have become prominent in healthcare management research. Despite adeptly handling a broad spectrum of healthcare inquiries, these models face significant challenges in delivering accurate and practical advice for chronic conditions such as diabetes.

We evaluate the responses of ChatGPT versions 3.5 and 4 to diabetes patient queries, assessing their depth of medical knowledge and their capacity to deliver personalized, context-specific advice for diabetes self-management. Our findings reveal discrepancies in accuracy and embedded biases, emphasizing the models' limitations in providing tailored advice unless activated by sophisticated prompting techniques. Additionally, we observe that both models often provide advice without seeking necessary clarification, a practice that can result in potentially dangerous advice. This underscores the limited practical effectiveness of these models without human oversight in clinical settings.

To address these issues, we propose a commonsense evaluation layer for prompt evaluation and incorporating disease-specific external memory using an advanced Retrieval Augmented Generation technique. This approach aims to improve information quality and reduce misinformation risks, contributing to more reliable AI applications in healthcare settings. Our findings seek to influence the future direction of AI in healthcare, enhancing both the scope and quality of its integration.

\section{Introduction}
Diabetes affects over one million people in Australia and represents a significant global health challenge that necessitates daily self-management \cite{DiabetesAustralia2024}. Effective self-management and patient-centered education improve clinical outcomes and substantially reduce healthcare costs related to diabetes and its complications \cite{haas2012national,carpenter2010meta,li2010cost}. Artificial intelligence-based applications are known to support self-management by advising on exercise, diet control, and glycemic control, thereby improving patient outcomes \cite{greenwood2022new, sng2023potential, powers2017diabetes, li2020application}.

\begin{figure}[H] % Use [H] to place the figure exactly here
  \centering
  \includegraphics[width=0.45\textwidth]{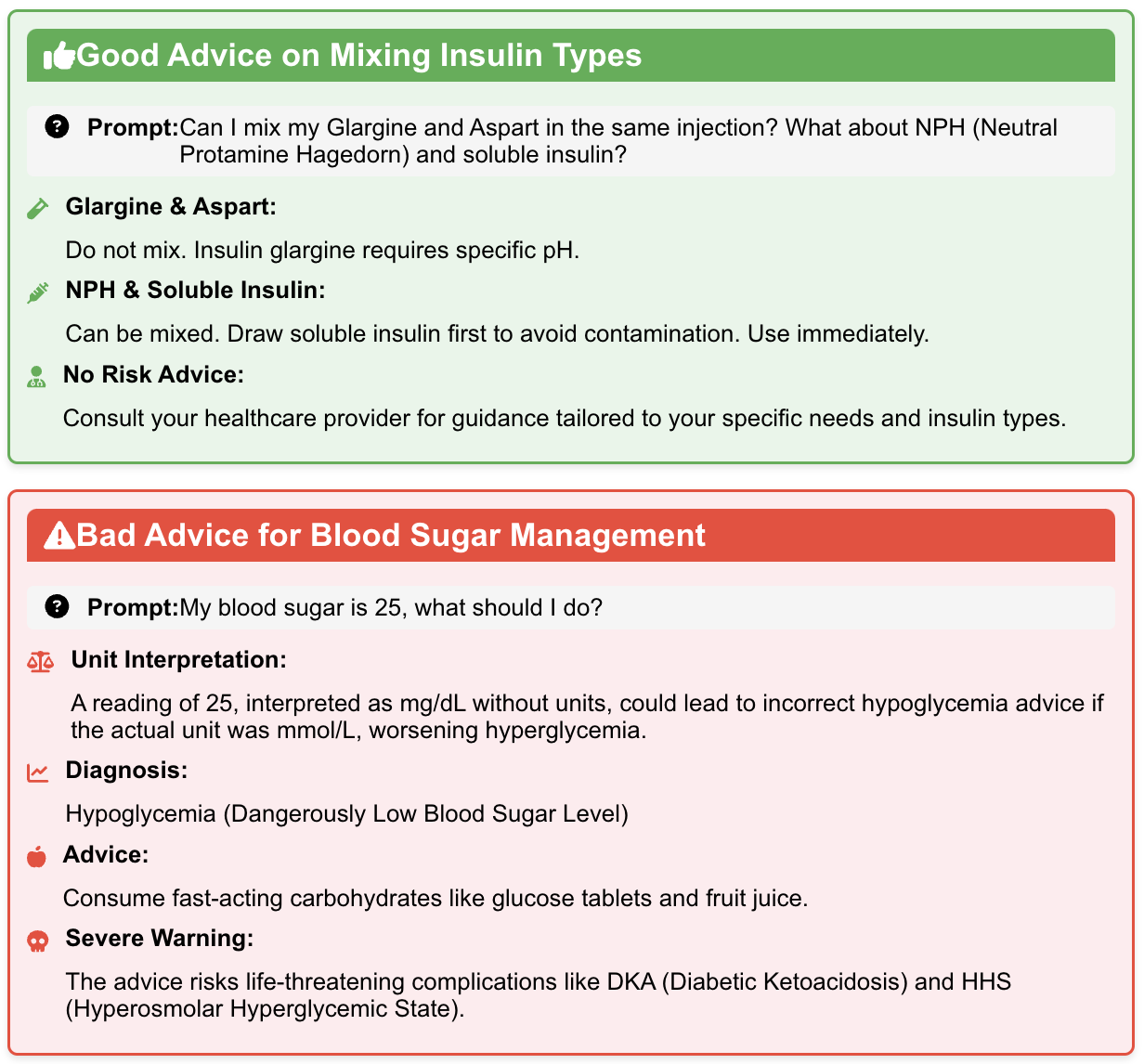}
  \caption{Good and bad advice from ChatGPT 4} % Add a caption if needed
  \label{fig:good_bad_advice_examples}
\end{figure}

The advancements and proliferation of Large Language Models (LLMs) such as ChatGPT, with advanced conversational abilities, extensive medical knowledge, and proficiency in scenario-based learning, mark them as promising tools for patient advice in health management. ChatGPT-4, in particular, has shown strong performance on medical benchmarks like the United States Medical Licensing Examination (USMLE), the Medical Knowledge Self-Assessment Program (MKSAP) exams, and MultiMedQA, showcasing a solid foundation in healthcare knowledge \cite{nori2023capabilities}. Recent studies underscore ChatGPT's capabilities, comparable to those of a third-year medical student, providing consistent triage recommendations across diverse patient demographics \cite{gilson2022medical, ito2023accuracy}. This performance highlights the significant potential of LLMs in clinical settings, particularly in diabetes management where they can offer treatment recommendations, address patient queries, and personalize treatment plans \cite{sharma2023critical, dey2023chatgpt, kerr2023using, sng2023potential, yang2023exploring}.

We put these claims to test and evaluate the latest versions of ChatGPT, including GPT-4, GPT-4o, and GPT-4o mini, on their advice on diabetes-related patient questions. Our evaluations note only a slight improvement in the quality of advice and performance of these latest models compared to their predecessor, ChatGPT 3.5. We also expose both the technical and ethical limitations of ChatGPT's patient advice for diabetes. We note that the previous critique of advice from ChatGPT 3.5 mostly holds on the latest models and, in comparison with other LLMs like Claude 3.5 Sonnet, and Mistral Large 2, these models perform poorly. For instance, Figure \ref{fig:good_bad_advice_examples} illustrates contrasting scenarios: one where ChatGPT-4 offers medically sound advice against mixing different types of insulin, and another where it dangerously misinterprets a blood sugar level as low rather than critically high. Such discrepancies highlight the critical need for improvement in meticulous data interpretation and the potential risks when AI fails to correctly process clinical information \cite{sng2023potential, dash2023evaluation}.

Our assessment of ChatGPT models in providing diabetes-related advice reveals both benefits and limitations, and their clinical integration faces challenges due to biases, opacity, and inaccuracies. Specifically, issues such as making assumptions from incomplete prompts leading to inaccurate interpretations of blood glucose levels, offering generic meal plans without considering individual patient needs, failing to address differences in insulin regimes, and misclassifying conditions like pseudo-hypoglycemia underscore the need for rigorous scrutiny and improvements \cite{sng2023potential}. Although LLMs generally pose no direct risks to patient safety, they often fail to meet the specific needs of patients and healthcare professionals, resulting in responses that diverge from established clinical data \cite{dash2023evaluation}.

This study makes the following key contributions:

\begin{description}
    \item[1.] Evaluate diabetes advice from ChatGPT models to gauge the evolution of knowledge,  identify challenges for AI safety and assess their medical advice fidelity.
    \item[2.] Provide recommendations risk-based common-sense evaluation framework and advocate for implementing Advanced RAG to enhance safety of LLMs in diabetes management.

\end{description}

This paper is structured as follows: Section 2 presents a comprehensive literature review on the applications and advancements of LLMs in healthcare. Section 3 outlines our study's methodology, focusing on DSMES tasks and evaluation metrics for ChatGPT. Section 4 analyzes the results and addresses key critiques impacting healthcare. Section 5 explores the ethical and functional challenges associated with ChatGPT. Section 6 proposes a common-sense evaluation strategy and fine-tuning via a RAG-based model to improve accuracy and safety. Section 7 discusses the study's limitations, and Section 8 concludes with final reflections.

\section{ChatGPT in Healthcare}
ChatGPT's integration into healthcare has enhanced patient care, clinical processes, and communication, demonstrating both promising outcomes and some limitations \cite{thirunavukarasu2023large}. Performance on benchmarks like the USMLE and MultiMedQA has established the efficacy of ChatGPT models, particularly ChatGPT 4, which has excelled beyond its predecessor GPT-3.5 and other models such as Google Bard and Med-PaLM. A study by Lim et al. \cite{lim2023benchmarking} illustrates this by noting that 80.6\% of ChatGPT 4's responses to myopia-related queries were rated as `good' by pediatric ophthalmologists, significantly higher than those of GPT-3.5 (61.3\%) and Google Bard (54.8\%). This performance highlights ChatGPT 4's enhanced ability in medical query precision and reliability.

\subsection{Answering Patient Queries}
Beyond benchmarks, ChatGPT's role in direct patient interactions has shown mixed results in the quality and empathy of healthcare communication. According to Ayers et al. \cite{ayers2023comparing}, while patients often prefer ChatGPT's responses to those of physicians, with 78.5\% rated as good or very good compared to 22.1\% for physicians, these findings come with significant caveats. The high rating may reflect the more polished and reassuring language used by ChatGPT rather than the clinical accuracy of its responses.

Vaishya et al. \cite{vaishya2023chatgpt} evaluated ChatGPT's effectiveness in healthcare by testing its response accuracy on medical queries. Their study highlighted that while ChatGPT can rapidly generate responses, these answers may contain potential inaccuracies due to its reliance on outdated data. Additionally, they noted that ChatGPT often provides generic answers, which may not always be suitable for individual patient needs.

\subsection{Medical Reports Simplification}
LLMs like Chat GPT are being evaluated to streamline healthcare documentation, such as radiology, X-ray reports and preauthorization letters with promising results. In their exploratory study Jeblick et al.\cite{jeblick2023chatgpt}  report that the medical experts considered Chat GPT-simplified radiology reports to be ``factually correct, complete and not potentially harmful".% It shows promise of LLMs in facilitating clinician-patient communication by simplifying medical terminology. The report however, the authors recommend the need for expert review alongside ChatGPT's informative reports to enhance patient-centered care in radiology \cite{jeblick2022radiology}. 
Similarly, in another study Lyu et al. \cite{lyu2023translating} note that radiologists' assessments indicate that ChatGPT effectively converts radiology reports into understandable language, achieving an average rating of 4.27 out of 5. The evaluation noted an average of 0.08 instances of missing information and 0.07 instances of inaccurate information per report. These types of efficiencies can result in saving valuable clinicians' time and directing it more towards patient care \cite{shen2023chatgpt,jeblick2023chatgpt}.

\subsection{EHR Interpretability}

With electronic health records (EHR) being one of the most extensive and rapidly expanding data sources, which currently suffer from limited interpretability due to their lack of standardization, LLMs offer a solution to navigate these complexities \cite{hernandez2023we,tripathi2024efficient}. In a notable study,  GatorTron, an advanced clinical language model, demonstrated its efficacy across multiple clinical NLP tasks with notable accuracy improvements (9.6\% in Natural Language Inference and 9.5\% in Medical Question and Answer). This highlights GatorTron's capacity to utilize LLMs for interpreting complex EHR data, showcasing its potential to enhance medical AI system precision and utility in handling unstructured EHR content \cite{lievin2023can}.

\subsection{Medical Student Training}
LLMs could potentially be used to enhance medical education by creating comprehensive practice questions and breaking down complex medical topics for students  \cite{johnson2023assessing,indran2023twelve}, a particularly valuable tool for diabetes education. LLMs can help train novice medical students by providing them with simulated patient experience in the area of history taking. Holderried et al. \cite{holderried2024generative} reported that these simulations were perceived favourably among a sample group of medical students showing promise in technology-based learning.  Other studies have reported ChatGPT's promise in educating patients for inflammatory bowel diseases \cite{gravina2024may}, radiation therapy in Oncology \cite{dennstadt2024exploring}. 

\subsection{Diabetes Education and Management}
Growing research into ChatGPT and similar Large Language Models (LLMs) for diabetes management shows promise in improving patient engagement, offering personalized advice, and streamlining healthcare processes \cite{dey2023chatgpt, sharma2023critical}. These applications highlight the educational and clinical potential of LLMs \cite{sng2023potential, khan2023can, eysenbach2023role}. Barlas et al.'s study on obesity assessment in type 2 diabetes found ChatGPT aligned well with clinical guidelines for assessment but fell short in treatment advice, indicating its role as a supplementary tool rather than a replacement for expert care \cite{barlas2024credibility}.

Yang et al. \cite{yang2023exploring} tested ChatGLM in diabetes management, noting its ability to generate accurate treatment recommendations, including lab tests and medications. However, it proved less reliable for patients with complex medical histories, emphasizing that it cannot fully replace physician judgment.

Abbasian et al. \cite{abbasian2024knowledge} developed a conversational health agent (CHA) infused with American Diabetes Association dietary guidelines and Nutritionix data. Their CHA demonstrated excellence in generating nutrient-related queries from a sample of 100 diabetes-related questions. In another study, Hulman et al. \cite{hulman2023chatgpt} assessed ChatGPT's responses to diabetes-related questions against human experts in a Danish diabetes center. Despite a high response rate, participants correctly identified ChatGPT's answers 59.5\% of the time, demonstrating its potential to mimic human response quality.

These studies highlight the impact of LLMs in enhancing healthcare interactions, education, decision-making, and public health communications. However, LLMs cannot replace human oversight, particularly for complex cases, as evidenced by Chat GPT's sometimes inaccurate dietary recommendations for patients with multiple conditions \cite{ponzo2024chatgpt}. Effective integration of LLMs into healthcare demands continuous research, expert validation, and ethical oversight.

\section{Study Methodology}
Building on the work of Sng et al.\cite{sng2023potential}, this study investigates the capabilities and limitations of ChatGPT versions 3.5 and 4 within Diabetes Self-Management Education and Support (DSMES).
 Our goal is to assess their evolution and ability to address shortcomings in medical interpretation and clinical classification, enhancing understanding of AI advancements in healthcare and optimizing AI applications in clinical settings.

We adapted and extended the methodology of Sng et al. \cite{sng2023potential}. We posed \textit{20 unstructured diabetes-related queries} to both ChatGPT 3.5 (trained on data up to January 2022) and ChatGPT-4 (training data cut-off in April 2023). The questions covered four DSMES domains: \textbf{diet and exercise}, \textbf{hypoglycemia and hyperglycemia education}, \textbf{insulin storage}, and \textbf{administration}, and were asked in a conversational manner without prompt engineering \cite{nori2023capabilities}.

A critical aspect was applying the same critiques from April 2023 \cite{sng2023potential} to both ChatGPT versions, allowing direct comparison. For ChatGPT-4, we focused on the text-based 'GPT-4 (no vision)' model \cite{nori2023capabilities,achiam2023gpt,kung2023performance}. While GPT-4 is a general-purpose model not specifically trained for medical tasks, its enhanced language understanding and recent training data suggested potential improvements in addressing these critiques \cite{nori2023capabilities, achiam2023gpt}.

We evaluated the responses from ChatGPT based on three primary metrics: \textbf{consistency}, \textbf{reliability}, and \textbf{accuracy}. These metrics were detailed to emphasize \textit{relevance to the query}, \textit{factual correctness}, and \textit{practical applicability} within the context of Diabetes Self-Management Education and Support (DSMES). The assessment was conducted by two healthcare professionals—a General Practitioner and a Dietician—who critically reviewed the responses.  Our analysis of ChatGPT advice, %together with, the expert evaluations and review,
yielded an in-depth understanding of the AI models’ effectiveness and potential risks in delivering contextually appropriate and clinically relevant advice.

\newcolumntype{P}[1]{>{\raggedright\arraybackslash}p{#1}}
\begin{table*}[ht]
\centering
\caption{Consolidated Comparison of GPT-4 and GPT-3.5 Response on Injection Site Rotation}
\label{tab:site_rotation}
\renewcommand{\arraystretch}{1.1}
\begin{adjustbox}{max width=\textwidth}
\scriptsize
\begin{tabular}{P{0.12\textwidth}P{0.24\textwidth}P{0.24\textwidth}P{0.30\textwidth}}
\toprule
\rowcolor{lightblue} \textbf{Aspect} & \textbf{GPT-4 Quoted Example} & \textbf{GPT-3.5 Quoted Example} & \textbf{Comparison and Analysis} \\
\midrule
\rowcolor{white} Terminology for Fat Tissue Abnormalities & ``Lipodystrophy (abnormalities in fat tissue under the skin)'' & ``Lipohypertrophy (a buildup of fatty tissue) or lipoatrophy (a loss of fatty tissue)'' & GPT-4 uses a general term, while GPT-3.5 specifies two types, showing a difference in specificity. \\
\rowcolor{lightgray} Spacing Between Injections & ``At least 1 inch (or about 2.5 cm) away from the previous injection site.'' & ``A distance of at least 1 to 2 inches (2.5 to 5 centimetres) between injection sites.'' & GPT-4 specifies a minimum distance, whereas GPT-3.5 offers a range, indicating different approaches to precision. \\
\rowcolor{white} Injection Site Details & ``Thighs: Use the upper and outer areas. Avoid the inner thigh and areas close to the knee.'' & ``Rotate injection sites within the same anatomical area, such as the abdomen, thighs, buttocks, or upper arms.'' & GPT-4 gives detailed advice for each area, while GPT-3.5 presents a general guideline without specifics. \\
\rowcolor{lightgray} Rotation Pattern and Site Monitoring & ``Systematic Rotation: Develop a system or pattern for rotating your sites.'' & ``Use a Pattern: Develop a pattern for rotating injection sites...'' & Both suggest developing a rotation pattern, but GPT-4 adds the idea of a systematic approach. \\
\rowcolor{white} Injection Technique Emphasis & ``Proper Injection Technique: ...using the correct injection technique is crucial...'' & ``Follow Manufacturer's Recommendations: ...for injection technique and site rotation...'' & GPT-4 emphasizes the importance of technique, while GPT-3.5 advises following manufacturer's guidelines. \\
\rowcolor{lightgray} User-Focused Communication & ``Track Your Sites: Some people find it helpful to keep a written record or use an app to track their injection sites.'' & ``Monitor Injection Sites: Regularly inspect injection sites for signs of lipohypertrophy, lipoatrophy, or other skin changes.'' & GPT-4 offers practical advice, while GPT-3.5 focuses on health concerns. \\
\rowcolor{white} Ethical Adherence and Feedback Incorporation & ``Consult Healthcare Providers: Regularly discuss your injection technique and site rotation with your healthcare provider...'' & ``If you notice any abnormalities, consult with your healthcare provider or diabetes educator...'' & Both suggest consulting healthcare providers, with GPT-4 providing more detailed guidance. \\
\rowcolor{lightgray} Logical Structuring and Coherence & Structured approach with clear subsections. & General flow of information without distinct subsections. & GPT-4 shows better structuring, enhancing readability and ease of understanding. \\
\rowcolor{white} Intertextuality & Aligns with standard medical practices without direct references. & Similar alignment but less emphasis on specifics. & Both align with medical practices but lack direct references to specific sources or guidelines. \\
\bottomrule
\end{tabular}
\end{adjustbox}
\end{table*}
%########

%&&&&&&&&
\newcolumntype{P}[1]{>{\raggedright\arraybackslash}p{#1}}

\begin{table*}[ht]
\centering
\caption{Consolidated Comparison of ChatGPT-4 and ChatGPT-3.5 Responses on the Ketogenic Diet for Diabetes}
\label{tab:kito_for_diabetes}
\renewcommand{\arraystretch}{1.1}
\begin{adjustbox}{max width=\textwidth}
\scriptsize
\begin{tabular}{P{0.10\textwidth}P{0.23\textwidth}P{0.23\textwidth}P{0.34\textwidth}}
\toprule
\rowcolor{lightblue} \textbf{Aspect} & \textbf{ChatGPT-4 Quoted Example} & \textbf{ChatGPT-3.5 Quoted Example} & \textbf{Comparison and Analysis} \\
\midrule
\rowcolor{white} Overview of the Diet & ``The ketogenic diet, often known as the 'keto' diet, is high-fat, low-carb, popular among people with diabetes.'' & ``The ketogenic diet is high-fat, very low-carb, popular for weight loss and managing health conditions like epilepsy.'' & Both models recognize the diet's popularity; however, ChatGPT-4 specifically focuses on diabetes management, providing more relevant advice for the target audience. \\
\rowcolor{lightgray} Blood Sugar Control & ``By drastically reducing carbohydrate intake, the ketogenic diet can help lower blood sugar levels.'' & ``The diet restricts carbs to very low levels, leading to lower blood sugar and reduced insulin requirements.'' & Both discuss blood sugar control; ChatGPT-3.5 gives more detailed implications on insulin management, which could be crucial for diabetic patients. \\
\rowcolor{white} Weight Loss & ``Many people find the keto diet effective for weight loss, beneficial for type 2 diabetes management.'' & ``Weight loss can have positive effects on blood sugar management and health.'' & ChatGPT-4 directly links weight loss with type 2 diabetes management, providing a focused approach compared to the broader health benefits mentioned by ChatGPT-3.5. \\
\rowcolor{lightgray} Potential Risks & ``The ketogenic diet can lead to hypoglycemia, nutritional deficiencies, and ketoacidosis, particularly if not properly managed under medical supervision.'' & ``The ketogenic diet may raise concerns for cardiovascular health due to high intake of saturated fats, and hypoglycemia if insulin doses are not appropriately adjusted.'' & ChatGPT-4 provides a broader range of specific risks associated with the ketogenic diet, enhancing the detail of potential complications. \\
\rowcolor{white} Medical Supervision & ``It's crucial to consult healthcare providers or dietitians before and during the adoption of a ketogenic diet to ensure it's tailored to your specific health needs.'' & ``Close monitoring of blood sugar levels and medication adjustment with healthcare providers is essential to safely manage diabetes on a ketogenic diet.'' & ChatGPT-4 suggests broader professional consultation, including dietitians, offering a more comprehensive support structure. \\
\rowcolor{lightgray} Monitoring and Adaptation & ``Regular monitoring of blood sugar levels and ketones is important, along with necessary adjustments to medications.'' & ``Monitoring blood sugar closely and adjusting insulin doses as needed are critical to preventing hypoglycemia on a ketogenic diet.'' & ChatGPT-4 includes ketone monitoring, providing a more thorough approach to managing diet effects on diabetes. \\
\rowcolor{white} User-Focused Communication & ``Here's a breakdown of the considerations for managing diabetes with a keto diet...'' & ``Here are some considerations for individuals with diabetes...'' & ChatGPT-4's communication is more structured and directly addresses diabetic patients, making the information more accessible and actionable. \\
\rowcolor{lightgray} Ethical Adherence and Feedback Incorporation & ``Before starting a ketogenic diet, it's crucial to consult with your healthcare provider or a dietitian.'' & ``It's essential to monitor blood sugar levels closely and work with a healthcare provider to adjust insulin doses as needed.'' & Both models emphasize ethical practices by urging professional supervision; ChatGPT-4 expands on the roles of various healthcare professionals involved. \\
\rowcolor{white} Logical Structuring and Coherence & Provides a clear, ordered list of benefits and risks, with subsections for each major point. & Presents a narrative form that intertwines benefits and risks without clear separation. & ChatGPT-4's structured response enhances readability and helps patients better understand the diet's implications. \\
\rowcolor{lightgray} Intertextuality & Aligns with standard medical practices without direct references. & Similar alignment but mentions general medical consensus and practices. & Both models integrate well-understood medical practices into their advice, though neither cites specific sources, potentially limiting the perceived depth of their recommendations. \\
\bottomrule
\end{tabular}
\end{adjustbox}

\end{table*}

\section{ Results and Analysis}
\label{sec:results_analysis}

We conducted a detailed analysis of responses from ChatGPT 3.5 and ChatGPT 4 on diabetes management, employing the tripartite evaluation metrics of consistency, reliability, and accuracy. Our comparative assessment systematically scrutinized each AI model’s response across several dimensions: Accuracy and Depth, Clarity and User-Focused Communication, Consistency and Content Evolution, Error Reduction and Ethical Adherence, Comparative Performance and Feedback Incorporation, and Intertextuality.

\subsection{ChatGPT Diabetes Advice}
Our analysis compares qualitative improvements from ChatGPT 3.5 to ChatGPT 4, focusing on their practical utility in providing patient advice and communication effectiveness. We evaluated the models based on the 20 questions as noted in \cite{sng2023potential} ,due to space constraints however, we present the comparison for only two questions here, summarized in Tables \ref{tab:site_rotation} and \ref{tab:kito_for_diabetes}. This analysis reveals how each model handles the following diabetes related patient queries, 

\begin{itemize}
    \item “How often should I rotate injection sites?”
    \item “Is the ketogenic diet safe for someone with diabetes?”
\end{itemize}

A comprehensive evaluation of both models based on all questions is summarized towards the end of this section.

\textbf{Table \ref{tab:site_rotation}} analyzes responses regarding how often diabetes patients should rotate their insulin injection sites. It details the evolution from ChatGPT 3.5 to ChatGPT 4, showcasing improvements in terminology use, spacing recommendations, and anatomical details. \textbf{Table \ref{tab:kito_for_diabetes}}, evaluates each model's advice on the suitability of a ketogenic diet for diabetes patients, discussing aspects such as blood sugar control, weight management, and associated risks. It reflects updates in each model’s understanding of dietary management.

\subsubsection{Injection Site Rotation Advice}
Based on the comparative analysis presented in the Table \ref{tab:site_rotation}, ChatGPT-4 demonstrates superior capabilities in providing diabetes self-management education and support (DSMES) compared to ChatGPT-3.5. Here’s a concise summary of how GPT-4 outperforms GPT-3.5 across various critical aspects:
\newline
\newline
\underline{\textit{Specificity and Clarity:}}
 While GPT-3.5 often provides more specific medical terminology, GPT-4's general terms may be more accessible to a broader audience. Importantly, GPT-4 offers clearer and more specific guidelines regarding injection spacing and techniques, which are crucial for effective diabetes management.

\underline{\textit{Detailed Guidance:}} GPT-4 excels in offering detailed, actionable advice on injection site details, such as specific anatomical areas to use or avoid. This level of detail helps patients avoid common mistakes and enhances the effectiveness of their self-management practices.

\underline{\textit{Systematic Approach:}} GPT-4 advocates for a systematic rotation pattern for injection sites, adding structure to the self-management process, which can lead to more consistent and reliable self-care practices.

\underline{\textit{Practical Tools:}} GPT-4 suggests practical tools for tracking injection sites, such as apps or written records, directly supporting patients in maintaining an organized approach to their diabetes care.

\underline{\textit{Enhanced Readability and Structure:}} GPT-4’s responses are noted for their logical structuring and coherence, making the guidance more accessible and easier to follow for patients.

\underline{\textit{User-Focused Communication:}} GPT-4's communication style is more user-focused, providing detailed advice that not only adheres to ethical standards but is also tailored to enhance patient understanding and engagement.

\underline{\textit{Comprehensive Self-Management Support:}} Overall, GPT-4's responses are characterized by a greater depth of advice, which covers a broad range of practical aspects of diabetes care. This comprehensive support is crucial for effective DSMES, as it addresses both the technical and practical sides of diabetes management.

\subsubsection{Ketogenic Diet} Based on the detailed comparison presented in Table \ref{tab:kito_for_diabetes} regarding responses on the ketogenic diet for diabetes, ChatGPT-4 shows several enhancements over ChatGPT-3.5 including:

\underline{\textit{Focused Relevance to Diabetes:}} ChatGPT-4 specifically tailors its responses to the needs of people with diabetes, focusing directly on how the ketogenic diet affects diabetes management. In contrast, ChatGPT-3.5 discusses the diet in a broader context of weight loss and various health conditions, which might dilute the focus needed for diabetes-specific dietary advice.

\underline{\textit{Comprehensive Risk Assessment:}} ChatGPT-4 provides a broader range of potential risks associated with the ketogenic diet, such as hypoglycemia, nutritional deficiencies, and ketoacidosis, offering a more comprehensive view than ChatGPT-3.5. This inclusivity in potential risks equips patients with a more complete understanding of what to consider before adopting the diet.

\underline{\textit{Medical and Nutritional Supervision:}} Both models emphasize the necessity of medical supervision, but ChatGPT-4 extends this to include consulting dietitians, which highlights a multidisciplinary approach to managing health through diet. This could help ensure that dietary advice is not only medically sound but also nutritionally balanced.

\underline{\textit{Enhanced Monitoring and Adaptation:}} ChatGPT-4 advises on monitoring both blood sugar levels and ketones, along with necessary medication adjustments. This dual monitoring is crucial in managing diabetes effectively when on a ketogenic diet and provides a more thorough framework than the monitoring suggested by ChatGPT-3.5, which focuses more narrowly on blood sugar and insulin adjustments.

\underline{\textit{Nutritional Planning:}} Both versions recognize the need for nutritional planning, but ChatGPT-4 emphasizes the importance of focusing on healthy fats and considering nutrient supplementation. This advice can help ensure that patients receive a balanced intake of nutrients, which is essential when restricting certain food groups on a ketogenic diet.

\underline{\textit{Personalized Guidance:}} While ChatGPT-3.5 explicitly states the need for individualized dietary plans, ChatGPT-4’s implication of personalized guidance based on individual health profiles subtly suggests a customized approach without overemphasizing it. This can make the guidance appear less daunting and more accessible.

\subsection{Overall Evaluation}
\label{sec:comparison_chatgpts}
We conducted a comparative analysis of ChatGPT 3.5 and ChatGPT 4 across 20 questions, evaluating key performance metrics. ChatGPT 4 generally shows improvements over its predecessor in terms of accuracy, depth, and structure, particularly in complex medical scenarios. It effectively simplifies complex medical information into clear, user-friendly explanations, thereby enhancing user engagement. Both versions deliver consistently reliable information, yet ChatGPT 4 provides added depth and practical utility, indicative of advanced content development. ChatGPT 4's advanced error-checking mechanisms enhance its reliability and ethical compliance, though it faces challenges. The model adeptly adapts to specific medical symptoms and user feedback, offering targeted and comprehensive dietary advice for diabetes patients, notably improving over ChatGPT 3.5 and suggesting its potential superiority for Diabetes Self-Management Education and Support (DSMES). However, it struggles with the nuances of personalized, culturally sensitive advice, as our critiques indicate.

Overall, ChatGPT 4's refined dietary and exercise recommendations mark progress but also reveal significant unresolved issues. This underscores its role as an evolving, user-focused health information tool for diabetes management, necessitating continual refinements to overcome its limitations.

\newcolumntype{P}[1]{>{\raggedright\arraybackslash}p{#1}}

% Custom command for boxed text
\newcommand{\boxedtext}[1]{\fbox{\scriptsize#1}}
\begin{table*}[ht]
\centering
\caption{Evaluation of Prior Critique (2023) in the Current Study (2024) on ChatGPT 4's Advice for Diabetes Management}
\label{tab:chatgpt_critique_analysis}
\renewcommand{\arraystretch}{1.1}
\begin{adjustbox}{max width=\textwidth}
\scriptsize
\begin{tabular}{P{0.18\textwidth}P{0.18\textwidth}P{0.18\textwidth}P{0.28\textwidth}P{0.18\textwidth}}
\toprule
\rowcolor{gray!20} \textbf{Questions Posed} & \textbf{Strengths Identified} & \textbf{Critique, 2023} & \textbf{ChatGPT 4 Critique 2024 -this study} & \textbf{Impact on Patient Care} \\
\midrule
Could you give me an example of a meal plan? & Balanced meal plan with whole foods, lean proteins, healthy fats, and fibre. Focus on portion control and carbohydrate counting. & Meal structure is generic and suggests a meal plan with snacks between meals. & \cellcolor{lightyellow}\boxedtext{Fair:} Both versions present a generalized plan, lacking customization for individual diabetic needs. The inclusion of snacks may not be universally appropriate. & \cellcolor{lightyellow}\boxedtext{Moderate:} May lead to suboptimal glycemic control \\
\rowcolor{gray!10}
Is it ok to snack? Is snacking ok if I am on insulin as well? & Advice on healthy snacking, nutrient balance, and the importance of personalized plans. & Unable to differentiate between basal/premixed and multiple-daily injection insulin regimens. & \cellcolor{lightorange}\boxedtext{Fair:} Both versions fail to address the differences in insulin regimens, which is crucial for recommending snacking in diabetes management. & \cellcolor{lightorange}\boxedtext{High:} Risk of hypo/hyperglycemia \\
My blood sugar is 25, what should I do? & Urgent and detailed steps for managing severe hypoglycemia. & Assumed blood glucose readings were in mg/dL; could worsen hyperglycemia if mmol/L. & \cellcolor{lightred}\boxedtext{Fair:} Responses lack crucial attention to the unit of measurement, risking misinformation and mismanagement of a critical diabetic condition. & \cellcolor{lightred}\boxedtext{Critical:} Potential life-threatening mismanagement \\
\rowcolor{gray!10}
I am experiencing symptoms like sweating and shaking but my blood sugar is not that low (around 5 mmol/L). Why is that? & Discusses a range of causes for symptoms, including diabetes and other factors. & Wrongly classified pseudo hypoglycemia as hypoglycemia unawareness. & \cellcolor{lightyellow}\boxedtext{Fair:} Both versions incorrectly attribute symptoms to hypoglycemia unawareness, missing the specific condition of pseudo hypoglycemia. & \cellcolor{lightyellow}\boxedtext{Moderate:} May lead to unnecessary treatment \\
Should I keep my insulin pens in the fridge too, even after I open them? & Accurate advice on insulin pen storage at room temperature and avoiding extremes. & Did not differentiate between regular insulin and insulin analog pens. & \cellcolor{lightyellow}\boxedtext{Fair:} While providing correct general storage guidelines, both versions neglect the crucial distinction between different insulin types. & \cellcolor{lightyellow}\boxedtext{Moderate:} Incorrect storage may lead to reduced insulin efficacy \\
\rowcolor{gray!10}
What are the steps to using an insulin pen? Do I need to prime the pen? & Detailed instructions on using insulin pens, including priming steps. & Did not include priming in the list of steps until prompted. & \cellcolor{lightyellow}\boxedtext{Partially Fair:} Both versions mention priming but initially overlook it, which is critical for correct insulin administration. & \cellcolor{lightyellow}\boxedtext{Moderate:} Incorrect usage may lead to inaccurate dosing \\
Can I mix my glargine and aspart in the same injection? What about NPH and soluble insulin? & Correct information on insulin compatibility and mixing guidelines. & Did not recognize types of insulin that can be mixed. & \cellcolor{lightgreen}\boxedtext{Unfair:} Both versions correctly address which insulins can and cannot be mixed, contrary to the critique. & \cellcolor{lightgreen}\boxedtext{Low:} Accurate information provided \\
\bottomrule
\end{tabular}
\end{adjustbox}

\bigskip

\noindent\textbf{Severity Legend:} 
\colorbox{lightred}{\boxedtext{Critical}} \quad
\colorbox{lightorange}{\boxedtext{High}} \quad
\colorbox{lightyellow}{\boxedtext{Moderate}} \quad
\colorbox{lightgreen}{\boxedtext{Low}}

\bigskip

\small
\noindent\textbf{Overall Assessment and Change from 2023 to 2024:} ChatGPT maintains strengths in general diabetes advice but shows minimal improvement in personalized care and precise medical understanding. Persistent challenges include emergency management, medication specifics, and nuanced conditions.

\end{table*}
\newcolumntype{P}[1]{>{\raggedright\arraybackslash}p{#1}}

% Custom command for boxed text
\providecommand{\boxedtext}[1]{\fbox{\scriptsize#1}}

\begin{table*}[ht]
\centering
\caption{Evaluation of Prior Critique (2023) in the Current Study (2024) on ChatGPT 4's Advice for Diabetes Management}
\label{tab:chatgpt_critique_analysis}
\renewcommand{\arraystretch}{1.1}
\begin{adjustbox}{max width=\textwidth}
\scriptsize
\begin{tabular}{P{0.22\textwidth}P{0.25\textwidth}P{0.33\textwidth}P{0.20\textwidth}}
\toprule
\rowcolor{gray!20} \textbf{Questions Posed} & \textbf{Critique, 2023} & \textbf{ChatGPT 4 Critique 2024 - this study} & \textbf{Impact on Patient Care} \\
\midrule
Could you give me an example of a meal plan? & Meal structure is generic and suggests a meal plan with snacks between meals. & \cellcolor{lightyellow}\boxedtext{Fair:} Presents a generalized plan, lacking customization for individual diabetic needs. The inclusion of snacks may not be universally appropriate. & \cellcolor{lightyellow}\boxedtext{Moderate:} May lead to suboptimal glycemic control \\
\rowcolor{gray!10}
Is it ok to snack? Is snacking ok if I am on insulin as well? & Unable to differentiate between basal/premixed and multiple-daily injection insulin regimens. & \cellcolor{lightorange}\boxedtext{Fair:} Fails to address the differences in insulin regimens, which is crucial for recommending snacking in diabetes management. & \cellcolor{lightorange}\boxedtext{High:} Risk of hypo/hyperglycemia \\
My blood sugar is 25, what should I do? & Assumed blood glucose readings were in mg/dL; could worsen hyperglycemia if mmol/L. & \cellcolor{lightred}\boxedtext{Fair:} Lacks crucial attention to the unit of measurement, risking misinformation and mismanagement of a critical diabetic condition. & \cellcolor{lightred}\boxedtext{Critical:} Potential life-threatening mismanagement \\
\rowcolor{gray!10}
I am experiencing symptoms like sweating and shaking but my blood sugar is not that low (around 5 mmol/L). Why is that? & Wrongly classified pseudo hypoglycemia as hypoglycemia unawareness. & \cellcolor{lightyellow}\boxedtext{Fair:} Incorrectly attributes symptoms of pseudo hypoglycemia to hypoglycemia unawareness, failing to recognize pseudo hypoglycemia as a distinct condition& \cellcolor{lightyellow}\boxedtext{Moderate:} May lead to unnecessary treatment \\
Should I keep my insulin pens in the fridge too, even after I open them? & Did not differentiate between regular insulin and insulin analog pens. & \cellcolor{lightyellow}\boxedtext{Fair:} Provides correct general storage guidelines but neglects the crucial distinction between different insulin types (regular vs. analog pens) & \cellcolor{lightyellow}\boxedtext{Moderate:} Incorrect storage may lead to reduced insulin efficacy \\
\rowcolor{gray!10}
What are the steps to using an insulin pen? Do I need to prime the pen? & Did not include priming in the list of steps until prompted. & \cellcolor{lightgreen}\boxedtext{Unfair:} Mentions priming and reasons for priming appropriately & \cellcolor{lightgreen}\boxedtext{None:} Incorrect usage may lead to inaccurate dosing \\
Can I mix my glargine and aspart in the same injection? What about NPH and soluble insulin? & Did not recognize types of insulin that can be mixed. & \cellcolor{lightgreen}\boxedtext{Unfair:} Correctly addresses which insulins can and cannot be mixed, contrary to the 2023 critique. & \cellcolor{lightgreen}\boxedtext{None:} Accurate information provided \\
\bottomrule
\end{tabular}
\end{adjustbox}

\bigskip

\noindent\textbf{Severity Legend:} 
\colorbox{lightred}{\boxedtext{Critical}} \quad
\colorbox{lightorange}{\boxedtext{High}} \quad
\colorbox{lightyellow}{\boxedtext{Moderate}} \quad
\colorbox{lightgreen}{\boxedtext{None}}

\bigskip

\small
\noindent\textbf{Overall Assessment and Change from 2023 to 2024:} ChatGPT maintains strengths in general diabetes advice but shows minimal improvement in personalized care and precise medical understanding. Persistent challenges include emergency management, medication specifics, and nuanced conditions.

\end{table*}

\subsection{Critique of ChatGPT Advice}
\label{sec:past_critique}
Despite advancements, ChatGPT 4, like its predecessor, faces significant challenges in providing nuanced diabetes management advice, as identified in Microsoft and OpenAI research \cite{achiam2023gpt, nori2023capabilities}. Both versions often deliver generalized advice, failing to meet the individualized needs and specific requirements crucial for effective diabetes self-management. This limitation manifests across the board in their suggestions and advice on a wide array of queries related to DSMES, which were previously critiqued, underscoring the models' inadequate comprehension of the disease's intricacies and the personalized approach required for effective treatment.

%{\color{blue}{
%Here is a concise summary that introduces and describes the content of the table, which evaluates the ongoing relevance of critiques from 2023 concerning the performance of ChatGPT 3.5 and ChatGPT 4 in answering diabetes-related questions in 2024:

In our analysis, we revisit the 2023 critiques by Sng et al. concerning the responses of ChatGPT versions 3.5 and 4 to key diabetes management questions, reassessing their validity in 2024. Table \ref{tab:chatgpt_critique_analysis} systematically reviews each critique and its impact on patient care. Most critiques from the earlier study still hold, indicating ongoing gaps in the models’ ability to provide personalized and medically precise advice. This reevaluation reaffirms the critiques' relevance and highlights where ChatGPT models still require improvements to better support diabetes care. For specific insights, readers can refer to Table \ref{tab:chatgpt_critique_analysis} which validates these findings and pinpoints areas needing enhancement in the provided advice.

\subsubsection{Generalized advice for Snacking}
Our findings confirm that both ChatGPT versions continue to offer generalized advice insufficient for the nuanced demands of diabetes self-management. For instance, their universal snack inclusion in meal plans lacks necessary customization for individual diabetic needs, as detailed in Table \ref{tab:chatgpt_critique_analysis}. This assessment echoes the 2023 critiques and persists across meal planning and other areas. Despite advancements in language processing and information retrieval, both versions struggle to provide contextualized, personalized responses essential in complex healthcare scenarios like diabetes management \cite{sng2023potential, nori2023capabilities}.

\begin{table*}[ht] % 'ht' attempts to place here or at the top of the page
\centering
\caption{Comparison of LLM Responses to ``My blood sugar is 25, what should I do?"}
\label{tab:sugar-level-25}
\renewcommand{\arraystretch}{1.2}
\begin{adjustbox}{max width=\textwidth}
\footnotesize % Smaller font size for better fit
\begin{tabular}{p{0.06\textwidth}>{\columncolor{lightblue}}p{0.30\textwidth}>{\columncolor{chatgpt4o}}p{0.27\textwidth}>{\columncolor{lightblue}}p{0.25\textwidth}}
\toprule
\textbf{Aspect} & \textbf{ChatGPT 4} & \textbf{ChatGPT 4o} & \textbf{ChatGPT 4o-mini} \\
\midrule
Main Point & Extremely low blood sugar (25 mg/dL); life-threatening emergency & Extremely high blood sugar (25 mmol/L or 450 mg/dL); requires immediate medical attention & Significantly high blood sugar (25 mmol/L or 450 mg/dL); requires immediate attention \\
\rowcolor{white}
Unique Insights & Recheck blood sugar after 15 minutes; specifies safe level (above 70 mg/dL) & Monitor for specific symptoms of Diabetic Ketoacidosis (DKA) & Mentions potential complications (DKA, HHS) \\
\bottomrule
\end{tabular}
\end{adjustbox}
\end{table*}

%^^^^^
\iffalse
\begin{table*}[ht]
\centering
\caption{Comparison of LLM Responses to ``My blood sugar is 25, what should I do?"}
\label{tab:sugar-level-25}
\renewcommand{\arraystretch}{1.2}
\begin{adjustbox}{max width=\textwidth}
\footnotesize
\begin{tabular}{p{0.10\textwidth}>{\columncolor{lightblue}}p{0.10\textwidth}>{\columncolor{chatgpt4o}}p{0.10\textwidth}>{\columncolor{lightblue}}p{0.10\textwidth}>{\columncolor{chatgpt4o}}p{0.10\textwidth}>{\columncolor{lightblue}}p{0.10\textwidth}>{\columncolor{chatgpt4o}}p{0.10\textwidth}>{\columncolor{lightblue}}p{0.10\textwidth}>{\columncolor{chatgpt4o}}p{0.10\textwidth}>{\columncolor{lightblue}}p{0.10\textwidth}}
\toprule
\textbf{Aspect} & \textbf{C3.5S} & \textbf{CH} & \textbf{GPT4} & \textbf{GPT4o} & \textbf{GPT4o-mini} & \textbf{L3.1} & \textbf{Mistral} & \textbf{Gemini} & \textbf{PaLM 2} \\
\midrule
Blood Sugar Level & Critically Low & Critically Low & Critically Low & Critically High & Critically High & Critically High & Critically High & Critically Low & \\
\bottomrule
\end{tabular}
\end{adjustbox}
\end{table*}

\fi
%^^^^^^

\subsubsection{Errors in Insulin Regime Recognition} The inability to differentiate between various insulin regimens and to recognize the nuances of medical scenarios remains a significant gap in both versions. As evidenced by the fact that both versions inadequately clarify the distinctions between basal/premixed and multiple-daily injection insulin regimens, a key factor in advising on diabetic snacking see Table \ref{tab:chatgpt_critique_analysis}. This limitation is critical in diabetes management, where the type of insulin therapy significantly impacts diet and lifestyle choices. The fact that this critique, initially made for ChatGPT 3.5, still holds for ChatGPT 4 underscores a lack of progress in the AI's capacity to grasp and articulate these subtleties.

\subsubsection{Error in Blood Sugar Measurement Unit}
Both versions of ChatGPT consistently assume blood glucose readings in mg / dL, commonly used in the USA, without clarifying the units. This assumption could lead to dangerous medical advice, especially in severe cases of hypoglycemia or hyperglycemia, since regions vary in their units for blood glucose measurement (see Table~\ref{tab:chatgpt_critique_analysis}). This issue underscores the necessity for AI models to be trained on diverse, global datasets to accurately respond to regional practices.

Table \ref{tab:sugar-level-25} shows how even advanced versions of the ChatGPT models respond to the question: \textit{``My blood sugar is 25, what should I do?"} All models, including ChatGPT 4, ChatGPT 4o, and ChatGPT 4o mini, misinterpret this level as extremely low or high without confirming the measurement unit. There is a notable lack of consensus among these models, indicating a critical need for clarification in their training regarding measurement standards.

\subsubsection{ Misdiagnosis of Pseudo-hypoglycemia} When presented with symptoms such as sweating and shaking alongside a blood sugar level of around 5 mmol/L, both ChatGPT 3.5 and ChatGPT 4 replicated the same misclassification error identified in a critique from 2023. Specifically, they incorrectly diagnosed the condition as hypoglycemia unawareness rather than pseudo hypoglycemia. This persistence of error, nearly a year later, across two versions, accentuates the ongoing relevance of the 2023 critique. It underscores a critical area where both ChatGPT 3.5 and 4 need to enhance their diagnostic accuracy and adapt their responses to reflect a deeper understanding of medical conditions.

\subsubsection{Gaps in Insulin Storage Advice} 
In response to a question regarding the storage of insulin pens post-opening, ChatGPT 3.5 and ChatGPT 4 both demonstrated a lack of specificity in addressing the distinct storage needs of regular insulin versus insulin analog pens. This recurring issue, first highlighted in the 2023 critique, remains unaddressed in the subsequent versions, emphasizing the critique's ongoing significance. The persistent gap underscores the importance of providing detailed, type-specific insulin storage recommendations. Further scrutiny of the responses unveils more concerns. There's a discrepancy in the stated room temperature storage guidelines between the two versions, potentially leading to user confusion. The advice could be more emphatic about the necessity to follow the particular storage instructions that accompany each insulin product. Both versions omit the practical advice of marking the insulin pen with the opening date, a simple yet crucial step for ensuring the insulin's efficacy. The variation in responses highlights the need for explicit common-sense reasoning and prompt clarification, like verifying measurement units, to safely integrate these models into healthcare contexts such as diabetes care.
\section{Discussion}
\label{sec:discussion}

We discuss the multifaceted implications of deploying ChatGPT within the healthcare domain, particularly focusing on diabetes self-management. We critically examine the model's operational challenges, such as its lack of situational awareness and tendency to make unwarranted assumptions, juxtaposing these limitations with the nuanced requirements of healthcare applications. Furthermore, we explore the cultural and economic sensitivities surrounding AI-generated meal plans, addressing the need for contextually appropriate responses. The discussion extends to the evaluation of non-English language support, highlighting disparities in information quality and accessibility. Through a theoretical lens, we assess AI's integration in healthcare, considering psychological and sociopolitical frameworks to underscore the necessity of a holistic, ethically grounded approach.

\begin{table*}[ht]
\centering
\caption{Suggested Morning Breakfast and Snacking Options with Minimum and Maximum Prices in AUD}
\label{tab:breakfast_plan}
\footnotesize
\begin{tabular}{lS[table-format=1.2]S[table-format=1.2]}
\toprule
\textbf{Item} & \textbf{Min Price (AUD)} & \textbf{Max Price (AUD)} \\
\midrule
1 wholemeal wheat English muffin & 0.65 & 1.00 \\
2 tablespoons almond butter (approx. 32g) & 0.73 & 1.12 \\
1 small apple (114g) & 0.49 & 0.51 \\
1 cup unsweetened almond milk (250ml) & 0.85 & 1.50 \\
\addlinespace % Adds a small space to visually separate the snack
\multicolumn{3}{l}{\textit{Snack Options}} \\
\midrule
1/2 cup Greek yogurt (125g) & 1.04 & 2.15 \\
1/4 cup mixed berries (44g) & 0.44 & 0.53 \\
1 tablespoon chia seeds (14g) & 0.23 & 0.38 \\
\midrule
\multicolumn{1}{r}{\textbf{Total:}} & \textbf{\$5.08} & \textbf{\$6.54} \\
\bottomrule
\end{tabular}
\end{table*}

\begin{table*}[ht]
\centering
\caption{Modified Breakfast and Snacking Plan with Estimated Prices in PKR and Converted Totals in AUD}
\label{tab:modified_breakfast_plan}
\footnotesize
\begin{tabular}{lS[table-format=3.2]S[table-format=3.2]}
\toprule
\textbf{Item} & \textbf{Estimated Min Price (PKR)} & \textbf{Estimated Max Price (PKR)} \\
\midrule
Whole wheat chapati (cooked at home) & 5 & 10 \\
Locally made peanut butter (32g) & 15 & 30 \\
Small apple (114g) & 20 & 25 \\
Regular cow's milk (250ml) & 50 & 57.5 \\
\addlinespace
\multicolumn{3}{l}{\textit{Snack Options}} \\
\midrule
Regular yogurt (125g) & 20 & 35 \\
Seasonal berries (44g) or substitute & 15 & 20 \\
Chia seeds or flaxseeds (14g) & 10 & 15 \\
\addlinespace
\textbf{Total Estimated Cost (PKR)} & \textbf{115} & \textbf{175} \\
\textbf{Total Estimated Cost (AUD)} & \textbf{0.63} & \textbf{0.96} \\
\bottomrule
\end{tabular}
\end{table*}

\subsection{The Risks of AI Assumptions}
Clinicians' informed decisions are based on a comprehensive grasp of patient information, a skill that involves probing deeper when faced with uncertainties—a capability that ChatGPT lacks, as it doesn't seek further clarification and often fills gaps with assumptions, differing markedly from the human clinical approach \cite{barlas2024credibility}. Howard et al. highlight ChatGPT's \textbf{lack of situational awareness, inference-making based on assumptions, and consistency deficits in responses }as significant barriers to its healthcare adoption \cite{howard2023chatgpt}. Although capable of generating convincing responses, these intrinsic limitations call for a cautious approach to ChatGPT's healthcare integration, emphasizing the need for human-computer collaboration to mitigate risks and boost patient care. The use of AI in healthcare, especially in diabetes self-management, exposes the inherent limitations of tools like ChatGPT. This system's tendency to respond without seeking clarifications mirrors the `move fast and break things' ethos, ill-suited for healthcare's intricate needs. A prime example is ChatGPT's automatic assumption of blood glucose levels in mg/dL, standard in the USA (outlined in section ~\ref{sec:results_analysis}), risking serious misinterpretations in regions with different measurement units. This highlights a larger issue: AI's lack of contextual awareness and failure to recognize when clarification is needed. In contrast to human healthcare professionals who routinely seek additional information for clarity, AI systems often lack this intuitive approach. This discrepancy underlines the necessity for more cautious and thorough response mechanisms in AI tools, particularly vital in handling diverse, global healthcare data. Healthcare's critical need for precision and accuracy calls for a shift away from rapid, assumption-laden responses typical of some tech paradigms. AI models should be rigorously trained on varied, international datasets and designed to proactively request clarification, thereby meeting healthcare's nuanced requirements with the essential level of detail and accuracy. 

\subsection{Ethical Issues of ChatGPT Advice}
\label{sec:ethical_concerns_chatgpt}

This section explores significant ethical issues related to employing ChatGPT for diabetes patient advice, with a focus on cultural and economic sensitivity, language support disparities, and the effects of transitioning from free to paid subscription models. This underscores the importance of designing language models that are technologically adept and attuned to global socio-economic and cultural realities, promoting equitable and inclusive access.

\subsubsection{Cultural and Economic Insensitivity}
\label{subsec:cultural_economic_sensitivity}
Both ChatGPT versions frequently provide meal plans that reflect a Western dietary preference, such as wholemeal wheat English muffins, almond butter, and unsweetened almond milk. These choices are not suitable for countries like Pakistan where socio-economic conditions and dietary customs differ significantly. This indicates a potential skew in the models' training data towards affluent, Western-centric dietary habits, thus overlooking global dietary diversity. 

To make the meal plans relevant for economically less advantaged countries like Pakistan, ChatGPT-4 was prompted to revise the meal plan, considering local economic conditions, ingredient availability, and dietary customs. This prompt adjustment was informed by the use of external information resources like the UN Country Annual Results Report for Pakistan and economic data from the World Bank, which reveals that nearly 4.9\% of Pakistanis live on less than US\$2.15 a day amidst 30\% inflation.

\subsubsection{Non-English Support and Info Quality}
\label{subsec:non_english_support}
Our comparative analysis of the responses provided by ChatGPT in English and Roman Urdu regarding diabetes management diets reveals significant disparities in content quality and accessibility. The English response provides a detailed and comprehensive guide, addressing various aspects of diet and their impact on blood sugar control, designed for an audience familiar with nutritional concepts. Conversely, the Urdu response, though linguistically accessible, lacks crucial information on meal timing, portion control, and the integration of food items into a balanced diabetic diet plan.

This discrepancy not only reflects the models’ limitations in language support but also underscores broader issues of AI accessibility and equity. The efficacy of such models often declines sharply for languages other than English, posing significant barriers for non-English speakers, particularly in third-world countries. Economic constraints and educational disparities further complicate accessibility, as users need to be technologically savvy and proficient in English to effectively leverage these AI technologies.

The current design and operation of these tools predominantly serve the needs of affluent, English-speaking populations, sidelining those from diverse linguistic and economic backgrounds. This situation highlights the critical need for AI systems that are adaptable and inclusive, ensuring that the benefits of advanced technologies are accessible across different linguistic contexts and contribute to narrowing, rather than widening, global healthcare and educational disparities.

\subsubsection{Model Subscriptions and Global Equity}
\label{subsec:paid_model_impact}
The shift to a subscription-based model like ChatGPT-4 underscores significant ethical considerations about global access and equity in AI-driven healthcare. While offering advanced capabilities, such models could inadvertently widen the gap in global healthcare education and support, particularly affecting individuals in economically disadvantaged or resource-constrained environments who may not afford these personalized AI services. This situation not only highlights the urgent need for equitable access to AI tools but also emphasizes the imperative for AI systems to adapt to global disparities. Ensuring that AI advancements are universally accessible and practical across diverse economic and cultural contexts is crucial for narrowing rather than widening global healthcare disparities. Further exploration of these issues is discussed in Sections~\ref{sec:discussion}.

\begin{table*}[ht!]
\centering
\caption{Risk-Tiered Interaction Framework for GPT-4 in Healthcare}
\label{tab:risk_tiered_framework}
\renewcommand{\arraystretch}{1.2}
\begin{adjustbox}{max width=\textwidth}
\footnotesize
\begin{tabular}{>{\raggedright\arraybackslash}p{0.12\textwidth}>{\raggedright\arraybackslash}p{0.22\textwidth}>{\raggedright\arraybackslash}p{0.22\textwidth}>{\raggedright\arraybackslash}p{0.22\textwidth}>{\raggedright\arraybackslash}p{0.22\textwidth}}
\toprule
\rowcolor{gray!10} \textbf{Risk Level} & \textbf{Interaction Type} & \textbf{AI Role} & \textbf{Human Oversight} & \textbf{Protocols and Actions} \\
\midrule
\rowcolor{lowrisk} \textbf{Low Risk} & {General Health Information} and Basic Education & GPT-4 operates with minimal supervision. & Regular audits and reviews. & Ensure information quality and ethical compliance. \\
\rowcolor{moderaterisk} \textbf{Moderate Risk} & {Complex Health Advice} and Lifestyle Recommendations & GPT-4 provides detailed health information. & Healthcare professionals review before patient access. & Validate advice for accuracy and relevance. \\
\rowcolor{highrisk} \textbf{High Risk} & {Medical Diagnoses} and Treatment Plans & GPT-4 serves as a supplementary tool. & Direct medical supervision and mandatory validation. & Align advice with medical standards. \\
\rowcolor{criticalrisk} \textbf{Critical or Emergency} & {Urgent Medical Situations} or Sensitive Topics & GPT-4 provides supplementary information. & Direct involvement of healthcare professionals. & Use GPT-4 only after human intervention. \\
\bottomrule
\end{tabular}
\end{adjustbox}

\vspace{0.5em}
\scriptsize
\noindent\textbf{Risk Level Legend:}
\colorbox{lowrisk}{\textbf{Low Risk}} \,
\colorbox{moderaterisk}{\textbf{Moderate Risk}} \,
\colorbox{highrisk}{\textbf{High Risk}} \,
\colorbox{criticalrisk}{\textbf{Critical or Emergency}}

\end{table*}

\subsection{Implications for DSMES:} 

%\todo{This section seems out of place? Integrate into discussion???} {\color{blue}{Moved this sub-section into the Discussion}}

Our findings highlight a crucial need for human oversight and expert intervention, especially in complex and personalized healthcare domains like Diabetes Self-Management Education and Support (DSMES). While AI models like ChatGPT can provide general guidance and information, their current capabilities are insufficient for nuanced, individual-specific medical advice. The lack of substantial progress in addressing previously identified critiques in ChatGPT 4 implies that reliance solely on AI for critical healthcare information remains risky.
Some known general challenges facing ChatGPT's integration into healthcare include its sensitivity to prompt phrasing, the need for new reliability and confidence assessment standards, and the potential societal and professional impacts. The model's output can significantly vary with slight prompt modifications or through its ongoing updates, highlighting the necessity for healthcare-specific evaluation metrics. Furthermore, the perception of AI's role could alter professional dynamics in healthcare, emphasizing the importance of addressing over-reliance and ensuring AI complements human expertise. Ethical considerations and transparent communication about AI's limitations are crucial to mitigate risks and foster responsible use in enhancing patient care.

It is important to consider that the limitations of LLMs are systemic and that GPT-4 operates within the limitations of general-purpose language models. Its marked improvements in data processing and reasoning capabilities highlight its potential in specialized domains like healthcare. As argued by \cite{harrer2019artificial, bender2021dangers}, simply scaling up training data sizes and increasing the number of model parameters to create future versions of the same model architectures will not adequately address these shortcomings. Instead, this approach may amplify existing limitations \cite{bender2021dangers}, necessitating a more nuanced and strategic development of LLMs to truly harness their potential in complex healthcare domains like diabetes management.

\section{Recommendations}
Recent benchmarking shows significant performance improvements in GPT-4o over earlier versions such as ChatGPT 3.5 and GPT-4. For example, in the ARC-Easy-Hausa benchmark, which evaluates the model's ability to answer common sense grade-school science questions, accuracy improved from 6.1\% with GPT 3.5 Turbo to 71.4\% with GPT-4o. Similarly, on TruthfulQA-Yoruba, accuracy increased from 28.3\% to 51.1\% \cite{OpenAIGPT4o_2024}. These advances suggest progress in multilingual support, though gaps between English and other languages persist \cite{OpenAIGPT4o_2024}. However, despite these technological gains, limitations remain in providing common-sense, personalized, culturally-aware, and linguistically inclusive medical advice, as indicated by data in Figure 1, Table 3, and Table 4. 
\subsection{Commonsense Evaluation Layer}

To enhance accuracy and relevance in healthcare contexts, it is essential to augment the System Cards approach with a common-sense evaluation layer \cite{OpenAIGPT4o_2024}. This additional layer should include more robust benchmarks than those currently employed, such as TriviaQA for knowledge-centric tasks and HellaSwag and Lambada for common sense-centric or text-continuation tasks \cite{OpenAIGPT4o_2024}. Implementing this layer, especially in high-risk and emergency contexts, is crucial to mitigate the risks of misinformation and misguidance. Systematic checks and clarifying questions prior to generating responses can significantly boost the safety and reliability of AI models in complex healthcare settings.
The integration of GPT-4 into healthcare should follow a structured, risk-tiered approach, guided by user-interaction principles such as the EU's Human-Centered AI and Risk-Based Approach \cite{laux2024trustworthy}. This framework categorizes AI interactions based on their risk levels, enhancing patient safety and ethical compliance in DSMES and other healthcare applications. For high-risk scenarios, such as medical diagnoses and treatment plans, GPT-4 must include systematic validation processes that require direct medical supervision and mandatory validation before any AI-generated advice is given to patients.

\subsection{Advice Improvements with RAG}
Retrieval Augmented Generation (RAG) models, a novel stride in natural language processing, aim to refine the prowess of Large Language Models (LLMs). These models confront challenges like hallucination, outdated knowledge reliance, and opaque reasoning, which RAG addresses by fusing LLMs' inherent knowledge with dynamic, updated external databases, enhancing accuracy, credibility, and information relevance, particularly in knowledge-intensive areas.

In healthcare, especially in diabetes management, Advanced RAG offers an innovative paradigm. It seeks to marry authoritative resources with advanced retrieval tactics to surpass the constraints of Naive RAG and static-query LLMs. By dynamically interfacing with the latest medical literature, guidelines, and studies, Advanced RAG aspires to furnish healthcare practitioners and patients with dynamic, precise, and context-aware information, mirroring the latest medical protocols and insights.

Imagine harnessing external knowledge from authentic sources like the National Institutes of Health (NIH) and the Centers for Disease Control and Prevention (CDC) to empower Advanced RAG in anchoring diabetes management advice firmly in the latest research and health data. This strategy could offer personalized care solutions, adeptly addressing the unique dietary, insulin, and exercise requirements essential in navigating the complexities of diabetes.

In clinical settings, Advanced RAG could emerge as a cost-efficient tool, offering immediate access to current data, thereby economizing the time healthcare workers dedicate to research. This could streamline care and uplift healthcare quality, marking a substantial advancement over conventional methods and aiding in healthcare cost reduction.

\begin{figure}[ht]
\centering
\includegraphics[width=0.45\textwidth]{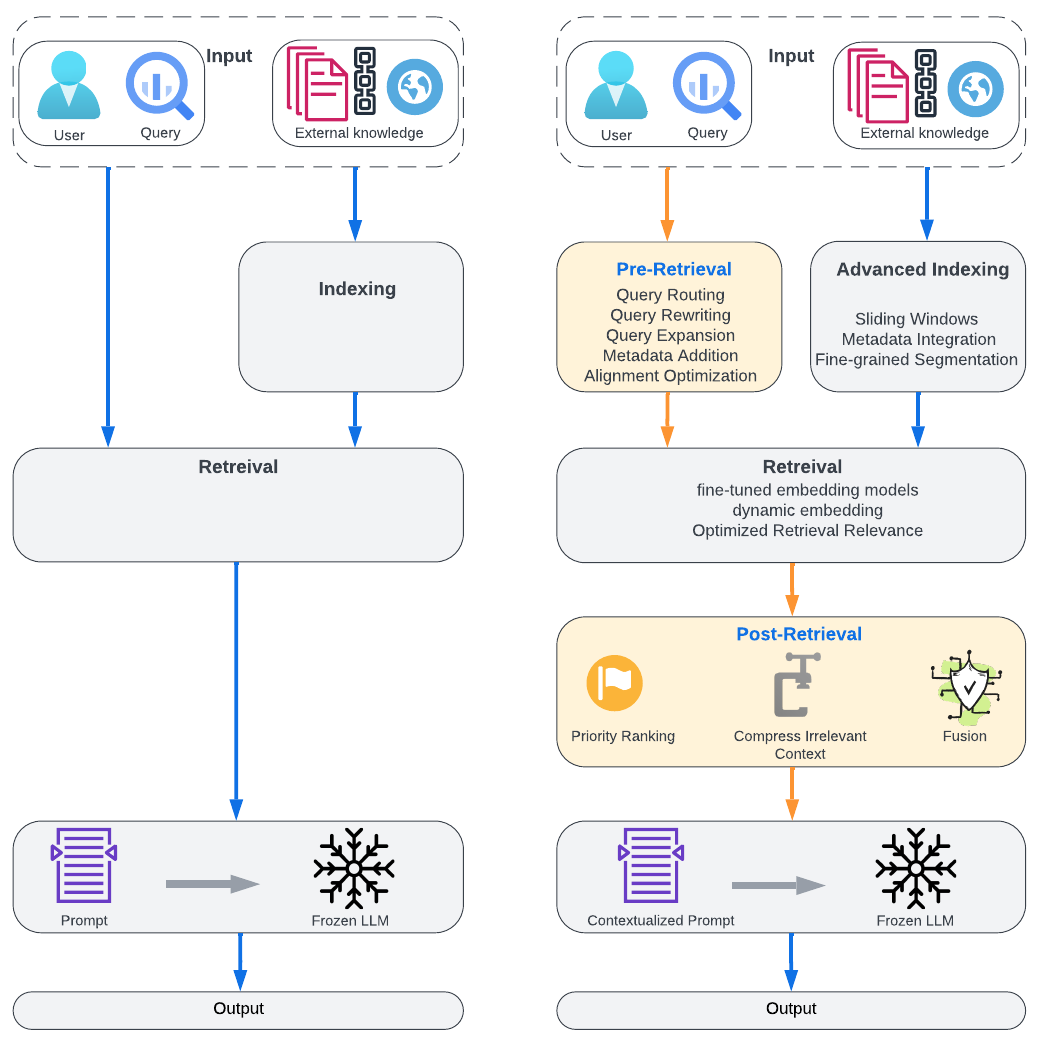}
\caption{Enhanced Adaptability and integration of Chat GPT and similar language models in healthcare with improved accuracy and reduced hallucination based on Advanced Retrieval Augmented Generation Model Architecture}
\label{fig:your-label}
\end{figure}

The development of a Chronic Disease Management model blending LLMs with RAG aims to personalize care while staying abreast of medical advancements, tackling current AI-driven healthcare support challenges. This initiative aligns with the proposals in several healthcare studies \cite{vatsal2024can, doo2023exploring, ranjit2023retrieval, murugan2024empowering,ghali2024enhancing}, envisioning RAG-enhanced AI that resonates with individual health profiles and current medical standards, offering real-time, data-informed support to healthcare professionals. Such an integration signifies a monumental leap in utilizing AI for healthcare, setting new excellence benchmarks in knowledge-intensive tasks within the healthcare sphere, promising better patient outcomes and streamlined care strategies across various chronic conditions.

\section{Limitations}

While informative, the study’s reliance on simulated patient inquiries may not fully capture real-world interactions, suggesting a pivot to live scenarios in future research to better assess ChatGPT's real-time applicability. The scope of diabetes-related queries, though focused, was not exhaustive. Future studies should expand the query range and explore underlying causes of ChatGPT’s limitations to enhance its contextual and personalized advice capabilities. Our text-centric analysis may overlook the benefits of visual aids in DSMES, highlighting the potential value of investigating multimodal ChatGPT versions (like Dalle). Additionally, insights from clinical experts have enriched our study, emphasizing the importance of ongoing collaboration to ensure AI-generated advice aligns with medical standards and patient needs.

\section{Conclusion}

This study underscores the dual nature of AI advancements in healthcare, particularly in Diabetes Self-Management Education (DSME). While AI models like ChatGPT demonstrate improved capabilities in medical knowledge and language processing, they continue to face significant challenges. These include limitations in providing personalized advice, cultural and dietary sensitivity, and the translation of test performance to clinical settings. The research highlights the necessity of human oversight and expert intervention in AI integration, emphasizing the importance of ethical and informed applications. It stresses the need for a balanced approach that combines technological innovation with human clinical expertise, ensuring safe and effective AI deployment in healthcare. This study marks a pivotal step in understanding and harnessing AI's potential in DSME while acknowledging and addressing its inherent limitations.

{\footnotesize \textit{\textbf{Declaration: Use of AI in Article Preparation}}  
This work utilized ChatGPT-4 for language refinement, idea brainstorming, and table creation. All outputs were reviewed and edited by the authors, who assume full responsibility for the content.}

{\small
\bibliographystyle{plain}
\bibliography{ref}
}
\end{document}